\documentclass[10pt, letterpaper]{article}
\usepackage{booktabs} 
\usepackage[english]{babel}
\usepackage[utf8x]{inputenc}
\usepackage[T1]{fontenc}
\usepackage[table,xcdraw]{xcolor}
\usepackage{multirow}

\usepackage[a4paper,top=3cm,bottom=2cm,left=3cm,right=3cm,marginparwidth=1.75cm]{geometry}

\usepackage{amsmath}
\usepackage{graphicx}
\usepackage[colorinlistoftodos]{todonotes}
\usepackage[colorlinks=true, allcolors=blue]{hyperref}
\usepackage{parskip}

\title{
		\usefont{OT1}{bch}{b}{n}
		\huge Self Expanding Convolutional Neural Networks \\
}
\usepackage{authblk}
\author[1]{Alex Deaconu}
\author[1]{Blaise Appolinary}
\author[1]{Sophia Yang}
\author[1]{Qingze (Eric) Li}

\affil[1]{University of British Columbia}
\date{January 10, 2024}

\begin{document}
\maketitle

\begin{abstract}\noindent
In this paper, we present a novel method for dynamically expanding Convolutional Neural Networks (CNNs) during training, aimed at meeting the increasing demand for efficient and sustainable deep learning models. Our approach, drawing from the seminal work on Self-Expanding Neural Networks (SENN), employs a natural expansion score as an expansion criteria to address the common issue of over-parameterization in deep convolutional neural networks, thereby ensuring that the model's complexity is finely tuned to the task's specific needs. A significant benefit of this method is its eco-friendly nature, as it obviates the necessity of training multiple models of different sizes. We employ a strategy where a single model is dynamically expanded, facilitating the extraction of checkpoints at various complexity levels, effectively reducing computational resource use and energy consumption while also expediting the development cycle by offering diverse model complexities from a single training session. We evaluate our method on the CIFAR-10 dataset and our experimental results validate this approach, demonstrating that dynamically adding layers not only maintains but also improves CNN performance, underscoring the effectiveness of our expansion criteria. This approach marks a considerable advancement in developing adaptive, scalable, and environmentally considerate neural network architectures, addressing key challenges in the field of deep learning.
\end{abstract} 


\section{Introduction}
Convolutional Neural Networks (CNNs) have revolutionized the field of deep learning, especially in processing grid-like data structures such as images [1]. Their effectiveness in tasks like image classification [2, 3], object detection [4, 5], semantic segmentation [6, 7] and image generation [8] stem from their ability to effectively learn spatial features. Convolutional layers, using filters or kernels, capture local patterns and extract features from input images. One important feature of convolutional layers is the shared weights implemented by kernels. This allows for efficient deep learning on images, as using only fully connected layers for such tasks would result in unfathomable numbers of parameters. Pooling layers, like max pooling and average pooling, reduce the spatial dimensions of these features, helping the network to focus on the most significant aspects.

Despite their popularity, CNNs face challenges in computational efficiency and adaptability. There have been several convolutional neural network architectures that have been proposed that are aimed at efficiency. Some of such architectures include MobileNet [12] and EfficientNet [13].  However, such traditional CNNs, with fixed architectures and number of parameters, may not perform uniformly across different types of input data with varying levels of complexity. 

Neural Architecture Search (NAS), a method for selecting optimal neural network architectures, has been a response to this challenge. NAS aims to obtain the best model for a specific task under certain constraints [14]. However, NAS is often resource-intensive due to the need to train multiple candidate models. in order to determine the optimal architecture. It is estimated that the carbon emission produced when using NAS to train a transformer model can amount to five times the lifetime carbon emissions of an average car [15]. This highlights the importance of finding suitable architectures for neural networks, yet also points to the limitations of current approaches in terms of static structure and proneness to over-parameterization.

Self Expanding Neural Networks (SENN), introduced in [9], offer a promising direction. Inspired by neurogenesis, SENN dynamically adds neurons and fully connected layers to the architecture during training using a natural expansion score (defined in section 2.1) as a criteria to guide this process. This helps overcome the problem of over-parametrization. However, its application has been limited to multilayer perceptrons, with extensions to more practical architectures like CNNs identified as a future research prospect.

Our study aims to develop a Self Expanding Convolutional Neural Network (SECNN), building on the concept of SENN and applying it to modern vision tasks. To the best of our knowledge, there has been no research on Self Expanding CNNs, despite the potential they hold for addressing model efficiency and adaptability in vision tasks. Unlike existing approaches that often require restarting training after modifications or rely on preset mechanisms for expansion, our approach utilizes the natural expansion score for dynamic and optimal model expansion. This research represents a significant step in developing adaptable, efficient CNN models for a variety of vision-related tasks.
The contributions of this research are as follows:
\begin{itemize}
\item Developing a Self Expanding CNN that dynamically determines the optimal model size based on the task, thereby enhancing efficiency.
\item Eliminating the need to train multiple CNN models of varying sizes by allowing for the extraction of checkpoints at diverse complexity levels.
\item Eliminating the need to restart the training process after expanding the CNN model.
\end{itemize}

\section{Methodology}
In order to develop a dynamically expanding convolutional neural-network architecture, we need an expansion criteria that triggers when to expand the model. The criteria we use is the natural expansion score as defined in section 2.1. 

\subsection{Natural Expansion Score}
The natural expansion score, as defined by [9], serves as a critical metric for assessing the effectiveness and uniqueness of augmenting the capacity of a neural network. This score is calculated by the inner product of the model's gradients and its natural gradients, encapsulated in the formula
\begin{flalign}
    && \eta = g^T F^{-1} g && 
\end{flalign}
\vspace{2mm}
Here, $F$ represents the Fisher Information Matrix (FIM), a key factor in understanding the loss landscape's curvature. The natural gradient, $F^{-1} g$, is an enhanced version of the standard gradient, adjusted to reflect the loss landscape's curvature. Recognizing the computational demands of calculating the full FIM, we utilize the Empirical Fisher as a practical approximation, focusing on the diagonal elements of the FIM. 
The empirical Fisher is given by
\begin{flalign}
&& F = \frac{1}{N} \sum_{i=1}^{N} \nabla_\theta L(\theta; x_i) \nabla_\theta L(\theta; x_i)^T &&
\end{flalign}
where: 
\begin{itemize}
    \item $L(\theta; x_i)$ is the loss function with respect to parameters $\theta$ for the $i$-th data point $x_i$
    \item $\nabla_\theta L(\theta; x_i)$ is the gradient of the loss function $L$ with respect to model parameters for the $i$-th data point $x_i$.
    \item $\nabla_\theta L(\theta; x_i) \nabla_\theta L(\theta; x_i)^T$ is the outer product of the gradient vector by itself
\end{itemize}
Intuitively, the natural expansion score $\eta$ captures the rate in loss reduction under a natural gradient descent [9].

Additionally, we add a regularization term to the computation of the natural expansion score to moderate the impact of adding new parameters to the network. This regularization is particularly important as it introduces a cost associated with increasing the model's complexity and prevents overexpansion. The modified formula for the natural expansion score, incorporating this regularization, is given by
\begin{flalign}
&& \eta_{\text{regularized}} = \eta \times \exp(-\lambda_n \times \Delta p^2) &&
\end{flalign}
where $\lambda_n$ represents the natural expansion score regularization coefficient and $\Delta p^2$ represents the increase of parameters of the proposed model from the previous one. The exponential decay factor, $\exp(-\lambda_n \times \Delta p^2)$, ensures that the score is significantly reduced as the number of new parameters from the addition increases. This regularization term is especially critical in preventing unnecessary or excessive growth of the network, aiding the expansion process to maintain an optimal balance between performance and complexity.

\subsubsection{Adding a New Layer}
When adding a new convolutional layer in the network, we ensure that the functionality of the current architecture is not affected by initializing the weights of the new layer as an Identity layer with a Gaussian noise. This strategy transmits all information from the preceding layer through to the new layer. This new layer adapts and learns its distinct weights later during training. Furthermore, the addition of the Identity layer ensures that there is a smooth integration of the layer into the network without a noticeable drop in model performance in the short run. This eliminates the need to restart training once we expand our model.

\section{Architecture Design, Initial Configuration, and Expansion Criteria}

Our neural network is built on a modular, block-based design, starting with a sequence of convolutional blocks. Each block initially consists of a single convolutional layer, a batch normalization layer, and a LeakyReLU activation function. This flexible design allows us to specify the number of channels for each block, supporting dynamic expandability.

\begin{figure}[h]
   \centering
   \includegraphics[width=1\textwidth]{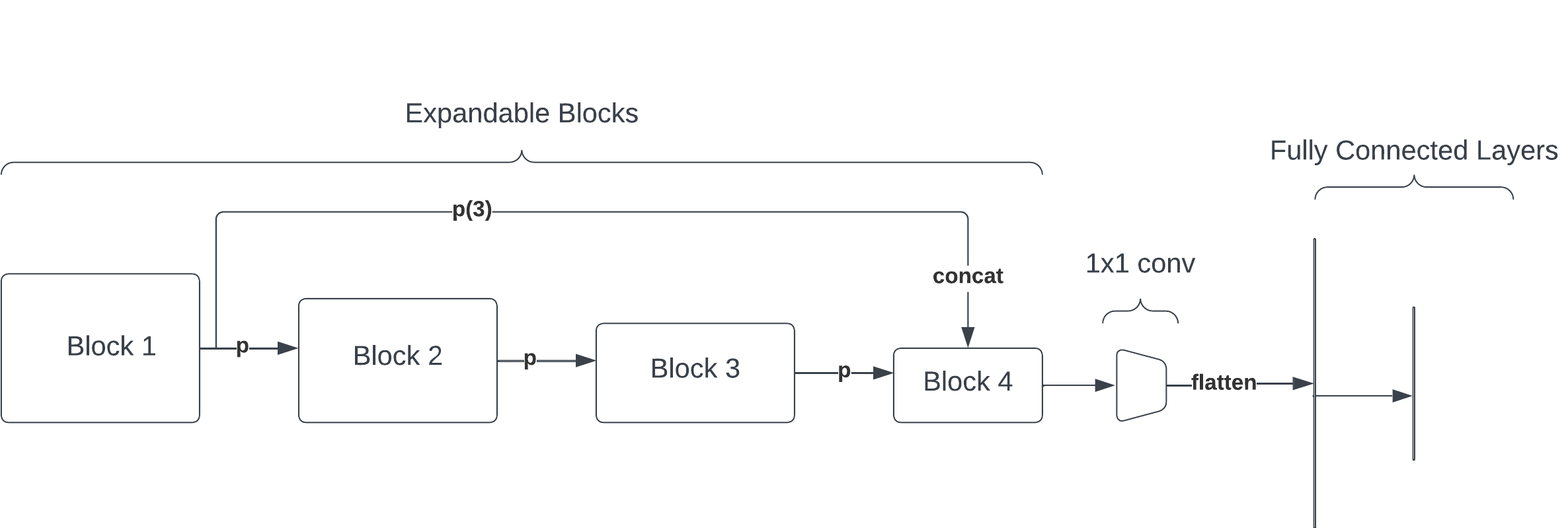}
   \caption{The model architecture. Each block includes a CNN layer, a batch normalization layer and a Leaky ReLU function. The blocks are separated by a pooling layer denoted by \textbf{p}. We include a skip connection from the first block to the output of the final block. We set a maximum capacity of each block to $N$. During training, the network dynamically expands by either adding an identity convolutional layer or upgrading the number of channels in a block, provided it does not exceed the block's capacity. These expansions occur when the network identifies a need for increased complexity to improve performance.}
   \label{The model architecture}
\end{figure}

The architecture processes input data through these blocks, engaging in feature extraction and refinement via layers of convolutions, normalizations, and activations. Pooling layers follow each convolutional block, crucial for reducing spatial dimensions and preventing overfitting.

The network concludes with a 1x1 convolutional layer with Leaky ReLU activation, a fully connected linear layer, and an output layer suitable for a 10-class classification dataset. Skip connections from the first to the final block are included to address the vanishing gradient problem.

\subsection{When, Where, and What to Expand}
We set a threshold $\tau$ that plays a significant role in the expansion. In determining when, where, and what to expand, our model employs the following steps:
\begin{itemize}
    \item Calculate the natural expansion score \(\eta_c\) of the current model.
    \item Create a temporary model with weights copied from the current model.
    \item For identity layer addition:
    \begin{itemize}
        \item For each block, add an identity convolutional layer.
        \item Track the block index leading to the highest natural expansion score \(\eta_l\).
    \end{itemize}
    \item For out channels increase:
    \begin{itemize}
        \item For each block, add \(C\) channels to every CNN layer, with Gaussian-distributed weights.
        \item Identify the block index yielding the highest expansion score \(\eta_i\).
    \end{itemize}
    \item Decide expansion type:
    \begin{itemize}
        \item If \(\eta_l > \eta_i\), and \(\frac{\eta_l}{\eta_c} > \tau\), add an identity layer to the original model at the optimal index.
        \item If \(\eta_i > \eta_l\), and \(\frac{\eta_i}{\eta_c} > \tau\), add channels to layers in the original model at the optimal index.
        \item If neither condition is met, the network does not expand at this moment.
    \end{itemize}
\end{itemize}
In our experimentation, we set the threshold $\tau =2$.

\subsection{Model Expansion Strategy and Implementation}

The expansion strategy of the model allows for the addition of layers or channels within set constraints. Each block can expand up to a maximum capacity \(N\) through either the addition of an identity convolutional layer or increasing the number of output channels.

The decision to expand is based on the natural expansion score, evaluated every epoch. Initially, channels are added in increments of 2, later adjusted to 4 to ensure thoughtful expansions.

Post-expansion, a 10-epoch cooldown period is implemented for network stabilization. When expanding a block, the output channels of all its layers are increased linearly, and the input channels of the next block's first layer are adjusted. New channels are initialized with Gaussian noise (mean=0) and a small coefficient (1e-4), balancing training progression with the integration of new parameters.

\subsection{Training}
\subsubsection{Training Dataset}
For evaluating our classification model, we used the CIFAR-10 dataset [11]. It comprises 60,000 colored images split 80-20 into training and testing sets. The data is divided into 10 mutually exclusive classes, with 6,000 images per class (5000 training, 1000 testing). Images are photo-realistic and only contain the class object, which can be partially occluded. CIFAR-10 is widely used and serves as a benchmark for evaluating model performance for image classification. See Figure 2 for a breakdown of the dataset classes with examples. The dataset was chosen for its large number of examples and its low resolution, which help minimize training time and computational resources
\begin{figure}[h]
   \centering
   \includegraphics[width=0.5\textwidth]{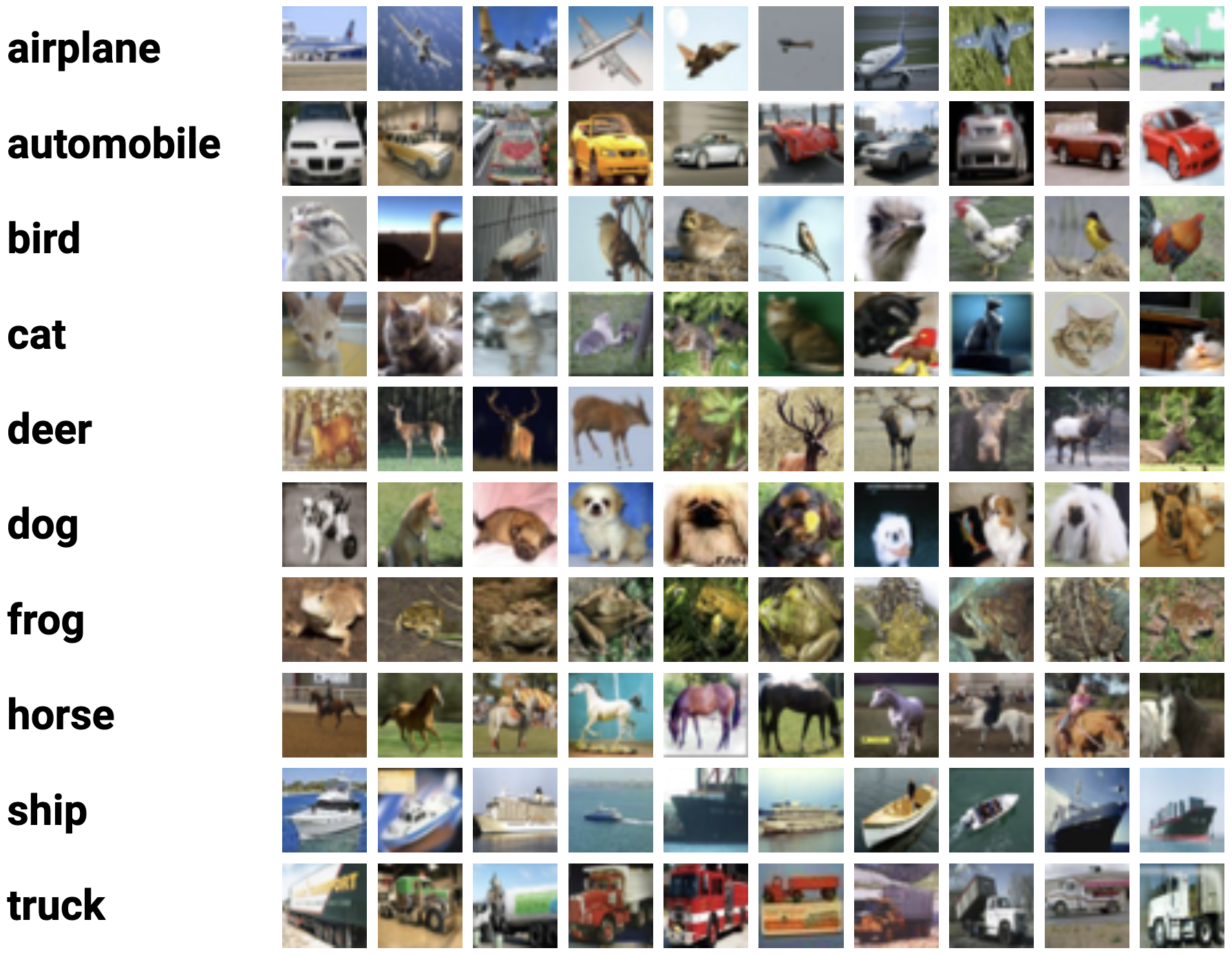}
   \caption{The 10 classes of the CIFAR-10 dataset along with classes [10].}
   \label{fig:yourlabel_}
\end{figure}

\subsubsection{Training Method}
We set our initial starting conditions to be a model consisting of 3 blocks, each with 16 output layers 3 input channels, on an NVIDIA GeForce RTX 3090 GPU, running on a Windows Desktop. 

We trained 5 models from identical starting conditions for 300 epochs, with a batch size of 512 on the CIFAR-10 dataset. We set the initial learning rate to 2e-3, which gets reduced by half after every 15 epochs without improvement in validation accuracy. To prevent overfitting we set our dropout rate to 0.1 for all convolutional layers, and 0.05 for the fully connected layers. This small dropout rate worked well, likely due to our model's small size which also prevented overfitting. To further combat overfitting, we used L1 regularized loss, with a regularization coefficient of 1e-5. We utilized minimal data augmentation in our training set, only using random  horizontal transforms to encourage the model to learn more robust features. Loss was calculated as cross entropy.
We set the maximum number of blocks per layer to be 10, to allow the model ample room for linear expansion while enforcing a limit to prevent overexpansion and prevent its associated problems such as the high computational cost and vanishing gradients.

\section{Results}
The results of our trials can be found in Figure 3 and a comparison of our model with other models on CIFAR-10 image classification can be found in Figure 4. 

\begin{figure}[h]
   \centering
   \begin{tabular}{lrrrrrrr}
    \toprule
     & & Number &  of & Parameters \\
    Metric & Trial 1 & Trial 2 & Trial 3 & Trial 4 & Trial 5 & Mean \\
    \midrule
    Val Accuracy (at 70\%) & 13696 & 11360 & 11360 & 11360 & 9024 & 11360.0 \\
    Val Accuracy (at 80\%) & 27852 & 51880 & 22636 & 37320 & 28588 & 33655.2 \\
    Highest Val Accuracy (\%) & 83.4 & 84.6 & 84.6 & 84.5 & 83.2 & 84.1 \\
    Parameters at Highest Accuracy & 74564 & 73720 & 57808 & 66440 & 40960 & 62698.4 \\
    \bottomrule
    \end{tabular}
    
   \caption{Table displaying the number of parameters required to achieve different validation accuracies on CIFAR-10 over 5 different trials with the same hyperparameters.}
   \label{fig:your__label}
\end{figure}
\begin{figure}[h]
   \centering
  \includegraphics[width=1\textwidth]{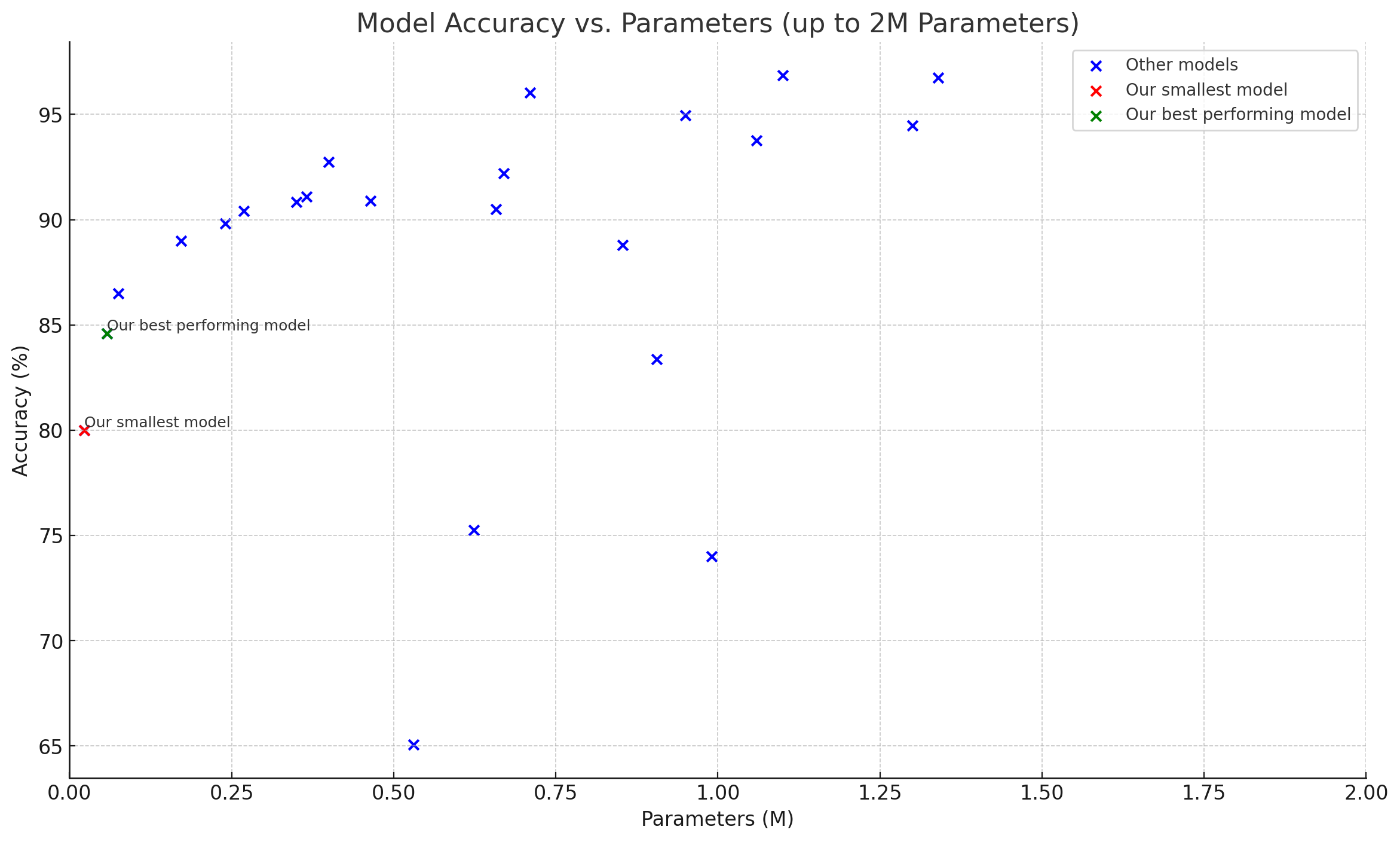}    
   \caption{Graph showing our best and smallest models compared to other models under 2M parameters [17].}
   \label{fig:your___label}
\end{figure}

\section{Discussion}
The trials on our Self-Expanding Convolution Neural Network (SECNN) on the CIFAR-10 dataset have yielded promising results, demonstrating our model's ability to dynamically grow and adjust to the complexity of the task. Our mean validation accuracy of 84.1\% is competitive considering our efficient approach to developing a full size model, as well as when we consider its simplistic approach. Its efficiency is further highlighted by the fact that this score was achieved with an average of 62,698 parameters, and an 80\% accuracy can be achieved with a mere 22,636 parameters. This is only 6\% lower than the existing most efficient architecture, ResNet-8 which is more complex architecture, features trainable activations, and over double the parameters [16]. Compared to the other models in Figure 4, SECNN is not only highly efficient but has a competitive validation accuracy as well.
Looking at our data, there is variation in results between trials. Due to our indeterministic expansion method, there is considerable variability in our results between trials. In some trials, the model opts to expand sooner and more aggressively, while in others, it follows a more gradual expansion pattern.
Because of our indetermistic expansion method, there is much variability in our results between trials, as in some trials the model decides to expand sooner and more aggressively, while other times the model decides expands in a more progressive pattern.
We approximated the Fisher Information Matrix as the Emprical Fisher Information Matrix, which is computationally efficient but less accurate than other approximations. Utilizing the Kronecker method [18] in our approach could potentially refine the calculation of the Fisher Information Matrix, leading to more precise natural gradients and possibly more effective expansions.
It is important that we address the limitations in our approach. For example, the CIFAR-10 dataset features very small images and may not have a full range of complexities and detail compared to real-world datasets. Therefore, our model's performance and behaviour as it expands may be different in more complex datasets, and may or may not be as effective.

\section{Conclusion}
In this article, we introduced the Self Expanding Convolutional Neural Network, a dynamically expanding architecture that uses the natural expansion score to optimize model growth. The CIFAR-10 dataset was used to train an initial model consisting over 5 different trials, which resulted in a 84.1\% mean validation accuracy. Our model demonstrates how a Self Expanding CNN offers a computationally efficient solution to dynamically determine an optimal architecture for vision tasks while eliminating the need to restart or train multiple models. 

\section*{Acknowledgements}
We would like to thank Rupert Mitchell, Martin Mundt and Kristian Kersting, whose work on Self Expanding Neural Networks greatly inspired our research.



\end{document}